\documentclass{article}
\usepackage{graphicx}
\usepackage{colortbl}
\usepackage{xcolor}
\usepackage{amsmath, amssymb}
\usepackage{parskip}
\usepackage{booktabs}
\usepackage{hyperref}
\usepackage{pdflscape}
\usepackage{multirow}
\usepackage{rotating}
\usepackage[table]{xcolor}
\usepackage{array}
\usepackage[margin=1in]{geometry}
\usepackage{adjustbox}
\usepackage{caption}
\usepackage{subcaption}
\usepackage{authblk}
\usepackage{chngcntr}
\usepackage{adjustbox}
\usepackage[normalem]{ulem}
\usepackage[table,dvipsnames]{xcolor}
\usepackage{booktabs}
\definecolor{maroon}{RGB}{128,10,10}
\usepackage{float}   
\usepackage{multirow}
\usepackage{array}
\usepackage{tikz}
\usetikzlibrary{shapes.geometric, arrows.meta, positioning, shadows.blur, calc}
\usepackage{xcolor}
\usepackage[normalem]{ulem}
\usepackage{tcolorbox}
\tcbuselibrary{skins}
\usepackage{dblfloatfix}
\usepackage{placeins}
\definecolor{graybg}{RGB}{249,250,251}
\definecolor{darkgray}{RGB}{55,65,81}
\definecolor{medgray}{RGB}{107,114,128}
\definecolor{lightgray}{RGB}{243,244,246}
\definecolor{bordergray}{RGB}{229,231,235}
\definecolor{finalbg}{RGB}{31,41,55}

\tcbset{
  compact/.style={
    before skip=6pt,
    after skip=6pt,
    top=4pt,
    bottom=4pt,
    left=4pt,
    right=4pt
  },
  phasebox/.style={
    enhanced,
    compact,
    colback=white,
    colframe=bordergray,
    arc=10pt,
    boxrule=1pt,
    drop fuzzy shadow,
    left=10mm,
    width=\textwidth
  },
  innerbox/.style={
    compact,
    colback=lightgray,
    boxrule=0pt,
    leftrule=4pt,
    arc=6pt
  }
}

\newcommand{\phasearrow}{%
\vspace{-2pt}
\begin{center}{\Large\color{medgray} ↓}\end{center}
\vspace{-6pt}
}

\title{Predicting Anemia Among Under-Five Children in Nepal Using Machine Learning and Deep Learning}
\author[1]{Deepak Bastola\thanks{Corresponding author: dbastola2022@fau.edu}}
\author[2]{Pitambar Acharya}
\author[3]{Dipak Dulal}
\author [4]{Rabina Dhakal}
\author[5]{Yang Li}
\affil[1, 5]{\small Florida Atlantic University, Department of Mathematics and Statistics}
\affil[2]{University of Alabama at Birmingham, Department of Applied Mathematics}
\affil[3]{Eastern New Mexico University, Department of Mathematical Science}
\affil[4]{Apara Innovations}
\date{}

\begin{document}
\maketitle
\section*{Abstract}
Childhood anemia remains a major public health challenge in Nepal and is associated with impaired growth, cognition, and increased morbidity. 
Using World Health Organization hemoglobin thresholds, we defined anemia status for children aged 6-59 months and formulated a binary classification task by grouping all anemia severities as \emph{anemic} versus \emph{not anemic}. 
We analyzed Nepal Demographic and Health Survey (NDHS 2022) microdata comprising 1,855 children and initially considered 48 candidate features spanning demographic, socioeconomic, maternal, and child health characteristics. 
To obtain a stable and substantiated feature set, we applied four features selection techniques (Chi-square, mutual information, point-biserial correlation, and Boruta) and prioritized features supported by multi-method consensus. 
Five features: child age, recent fever, household size, maternal anemia, and parasite deworming were consistently selected by all methods, while amenorrhea, ethnicity indicators, and provinces were frequently retained. 
We then compared eight traditional machine learning classifiers (LR, KNN, DT, RF, XGBoost, SVM, NB, LDA) with two deep learning models (DNN and TabNet) using standard evaluation metrics, emphasizing F1-score and recall due to class imbalance. 
Among all models, logistic regression attained the best recall (0.701) and the highest F1-score (0.649), while DNN achieved the highest accuracy (0.709), and SVM yielded the strongest discrimination with the highest AUC (0.736). 
Overall, the results indicate that both machine learning and deep learning models can provide competitive anemia prediction and the interpretable features such as child age, infection proxy, maternal anemia, and deworming history are central for risk stratification and public health screening in Nepal.\\

\textbf{Keywords:} childhood anemia, NDHS, machine learning, deep learning, Nepal

\section*{Introduction}
Anemia is characterized by a reduced number of red blood cells or a decreased hemoglobin concentration in the blood cells. According to WHO classification, anemia severity is categorized based on hemoglobin levels: no anemia ($\geq 11.0 $ g/dL), mild anemia (10.0-10.9 g/dL), moderate anemia (7.0-9.9 g/dL), severe anemia (<7.0 g/dL) \cite{WHOhemoglobin}. The global estimates show that 30\% of women aged 15-49, 37\% of all pregnant women, and 40\% of children aged 6-59 months are affected by anemia \cite{WHOhemoglobin}. The global burden of disease study estimated that anemia affected 1.9 billion people worldwide by 2021, with its most profound effects on children under 5, women, and countries in sub-Saharan Africa and South Asia, among others. Children under 5 years of age were predominantly affected, with the prevalence of 41.4\% worldwide. The prevalence of anemia was 43\% in South Asia that year \cite{gardner2023prevalence}. The consequences of anemia during childhood include impaired cognitive and motor development, poor nutrition, delayed growth, and increased morbidity, underscoring the importance of early interventions \cite{WHOhemoglobin}.

In Nepal, childhood anemia remains a significant public health problem. According to the Nepal Demographic and Health Survey (NDHS) 2022, about 43\% of children under 5 years old are anemic. \cite{NepalDHS2022_SR275} Historical data further highlight a worrying trend: the percentage of anemic children aged 6-59 increased from 46.4\% in 2011 to 52.2\% in 2016 \cite{NDHS2011, NDHS2016}. Though 2022 noted a moderate improvement in the anemia rate among this age group, the prevalence remains high and poses a major public health challenge. The burden of childhood anemia is even more notable among specific populations as it disproportionately affects poorer households, marginalized ethnic groups and certain ecological regions of Nepal \cite{sharma2022spatial}. Likewise, maternal factors such as maternal anemia, undernutrition during pregnancy, limited antenatal care, and birth order are strongly correlated with childhood anemia \cite{Upadhyay2025_anemiaBayesian, Chowdhury2020,  kuniyoshi2025maternal}. The same studies noted that children of younger age, those who are malnourished, with iron deficiency and helminth infection, residing in the Terai region, and from the Dalit ethnic group  are at a higher risk of anemia. Geographic disparities are also pronounced, with anemia prevalence markedly higher in the Terai plains than in the hilly and mountainous regions, differences often attributed to food insecurity, climatic variation, and uneven distribution of health resources \cite{harding2018determinants}. The availability of high-quality, nationally representative datasets, such as the NDHS, offers an unprecedented opportunity to examine anemia at scale. The NDHS provides a comprehensive collection of demographic, socioeconomic, and health indicators that can be utilized to understand population-level patterns and determinants of anemia. Previous research  has predominantly employed conventional statistical methods to examine the prevalence and associated factors of anemia in Nepal \cite{ kuniyoshi2025maternal, dhakal2022review}. However, with advances in computational power data science, machine learning (ML) and deep learning (DL) approaches now offer robust alternatives capable of modeling complex, nonlinear relationships in the demographic survey data \cite{mengistu2025understanding}. These methods enable high-dimensional analysis to capture hidden interactions among features.

Despite extensive epidemiological research on childhood anemia in Nepal \cite{sharma2022spatial,kuniyoshi2025maternal}, no studies to date have employed machine learning or deep learning methods for predictive modeling using NDHS data. This study addresses this critical gap by systematically benchmarking ten ML and DL algorithms to predict anemia status among Nepalese children aged 6-59 months. By rigorously comparing model performance across traditional classifiers and neural network architectures using stratified cross-validation and addressing class imbalance through SMOTE, we identify the optimal computational approach for community-based anemia screening in resource-limited settings. Our framework shifts the analytical paradigm from purely explanatory associations to actionable risk prediction, demonstrating that sociodemographic variables routinely collected in health surveys can effectively stratify children by anemia risk. The findings provide a foundation for developing practical screening tools that could inform targeted maternal-child health interventions to reduce anemia prevalence and its developmental consequences in Nepal.

\section*{Dataset}
\subsection*{Data Preprocessing}
Data preprocessing involved removing unwanted entries, dropping observations with missing target values, and excluding six high-missing variables (infant formula, juice intake, stool disposal, postnatal checkup, first breastfeed duration, vitamin A). Antenatal care was capped at 10 visits with median imputation; breastfeeding, iron during pregnancy, and mother deworming were imputed using mode. A binary malnutrition variable was created from anthropometric indicators, classifying children as nourished when all z-scores fell between -2 and +2 per WHO standards \cite{WHO2006}. After preprocessing, 1,855 observations with 13 categorical features were retained from the original 5,372 entries. This study converts all the features into categorical variables. During the model implementation, this study split data into 80-20 train-test with stratified sampling and employed Synthetic Minority Over-sampling Technique (SMOTE) to the training data, integrating into the pipeline to avoid data leakage, and used repeated stratified 5-fold cross-validation with 3 repeats. 

\subsubsection*{Feature Selection}
The response variable is childhood anemia, defined by hemoglobin levels according to WHO thresholds for children.  This study categorizes any stage of anemia as anemic and is designed for binary classification into anemic and non-anemic categories, involving 1,085 and 770 children, respectively. Firstly, 48 possible features were selected on the basis of similar research and employed four distinct statistical methods: Chi-Square, Mutual Information, Point-Biserial correlation, and the Boruta algorithm to robustly identify the most informative features of anemia status for NDHS data from the candidate set \cite{Zemariam2024}. 

\begin{figure}
    \centering
    \includegraphics[width=0.99\linewidth]{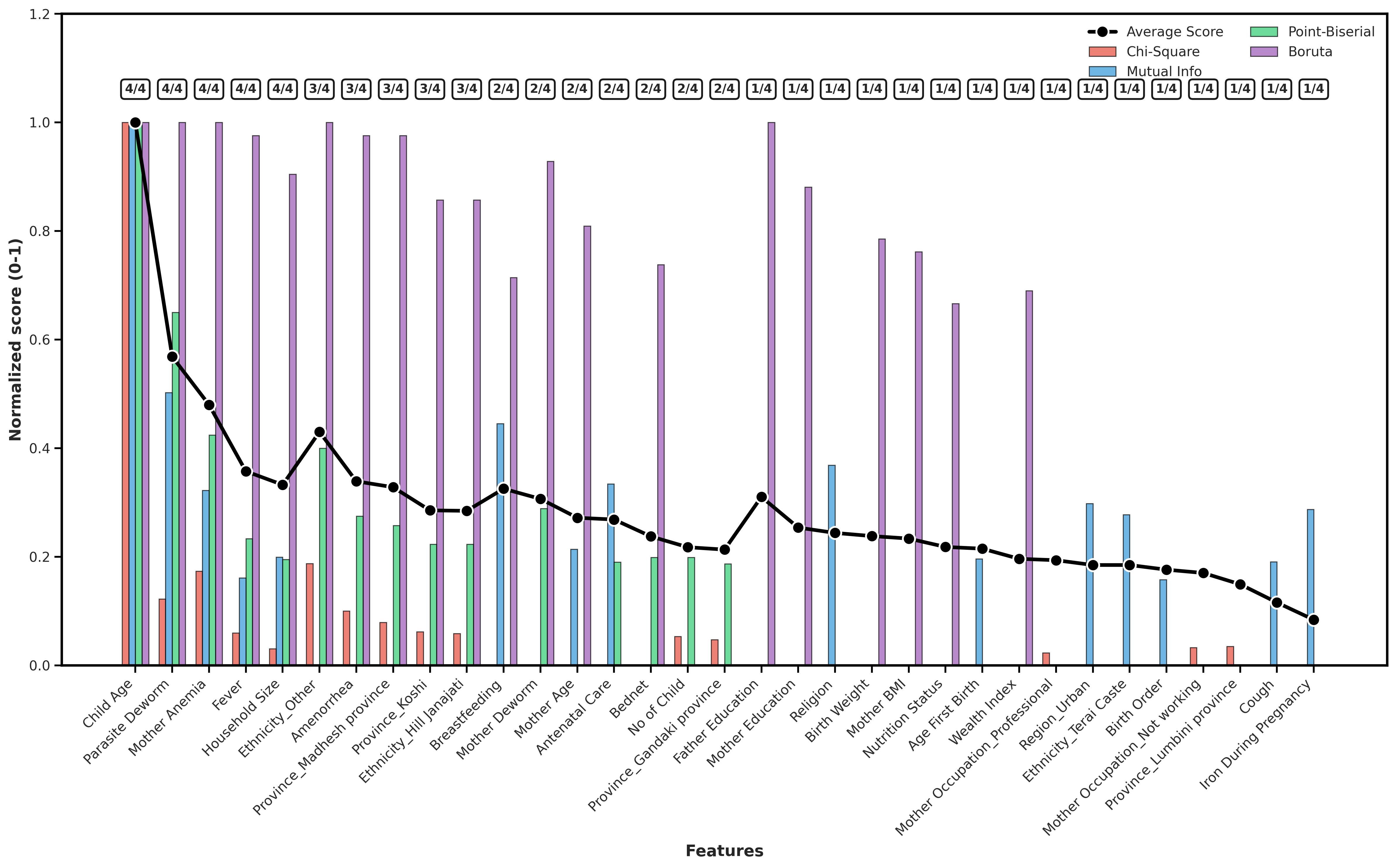}
    \caption{Normalized features importance across methods}
    \label{fig:consensus_features}
\end{figure}

Figure~\ref{fig:consensus_features} shows five features unanimously selected by all four methods: child age, fever, household size, mother anemia, and parasite deworming, indicating strong association with childhood anemia. Five additional features: amenorrhea, ethnicity (hill janajati, other), and provincial indicators (koshi, madhesh) were selected by three methods. Variables selected by two methods, including antenatal care, breastfeeding, and mother deworming, were retained based on their established significance in recent NDHS studies \cite{Upadhyay2025_anemiaBayesian,kuniyoshi2025maternal}. This multi-method consensus approach mitigates single-technique bias and enhances confidence in the final feature set.

\tcbset{
  phasebox/.style={
    enhanced,
    colback=white,                 
    colframe=darkgray,
    boxrule=0.7pt,
    arc=6pt,
    left=4pt,right=4pt,top=4pt,bottom=4pt,
    drop shadow={black!10!white}   
  },
  innerbox/.style={
    enhanced,
    colback=white,                 
    colframe=medgray,
    boxrule=0.5pt,
    arc=4pt,
    left=3pt,right=3pt,top=3pt,bottom=3pt,
    drop shadow={black!8!white}
  },
  compact/.style={
    enhanced,
    colback=white,                 
    colframe=darkgray,
    boxrule=0pt,
    toprule=3pt,
    arc=6pt,
    left=2pt,right=2pt,top=2pt,bottom=2pt,
    drop shadow={black!8!white}
  }
}

\providecommand{\phasearrow}{\par\noindent\centering{\color{medgray}\rule{0.35\linewidth}{0.4pt}}\par}
\FloatBarrier
\begin{figure*}[!t]
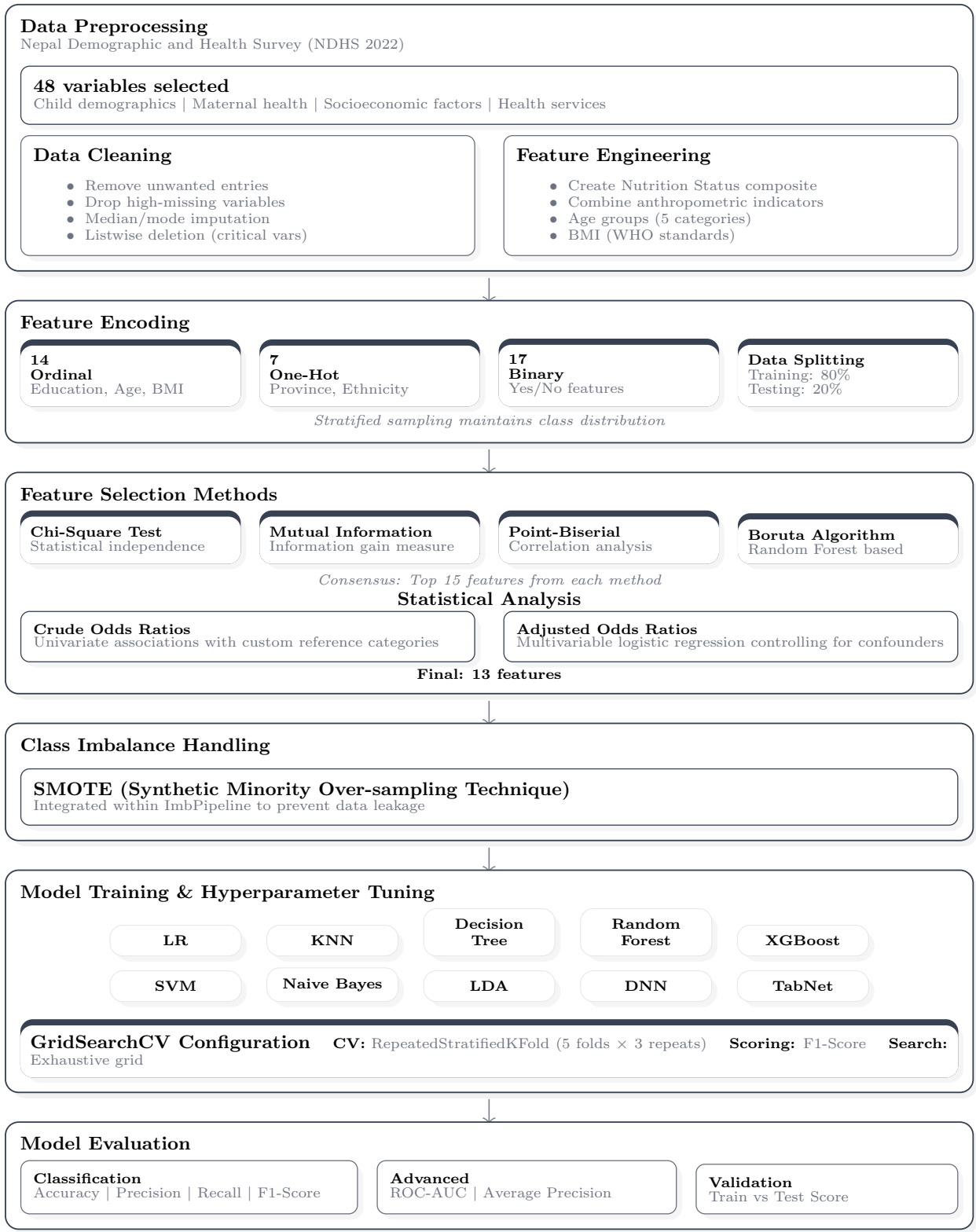

\centering

\resizebox{\textwidth}{!}{%
\begin{minipage}{\textwidth}
\scriptsize

\begin{tcolorbox}[phasebox]
{\bfseries \small Data Preprocessing}\\[-0.2pt]
{\scriptsize\color{medgray} Nepal Demographic and Health Survey (NDHS 2022)}\\[-6pt]

\begin{tcolorbox}[innerbox, colframe=darkgray]
{\scriptsize\bfseries \small 48 variables selected}\\
{\scriptsize\color{medgray} Child demographics $\mid$ Maternal health $\mid$ Socioeconomic factors $\mid$ Health services}
\end{tcolorbox}

\noindent\begin{minipage}[t]{0.485\textwidth}
\begin{tcolorbox}[innerbox, colframe=medgray]
{\bfseries \small Data Cleaning}\\[-2pt]
{\color{medgray}
\begin{itemize}
\item Remove unwanted entries
\item Drop high-missing variables
\item Median/mode imputation
\item Listwise deletion (critical vars)
\end{itemize}}
\end{tcolorbox}
\end{minipage}\hfill
\begin{minipage}[t]{0.485\textwidth}
\begin{tcolorbox}[innerbox, colframe=medgray]
{\bfseries \small Feature Engineering}\\[-2pt]
{\color{medgray}
\begin{itemize}
\item Create Nutrition Status composite
\item Combine anthropometric indicators
\item Age groups (5 categories)
\item BMI (WHO standards)
\end{itemize}}
\end{tcolorbox}
\end{minipage}
\end{tcolorbox}

\phasearrow

\begin{tcolorbox}[phasebox]
{\bfseries \small Feature Encoding}\\[-4pt]

\noindent
\begin{minipage}[t]{0.235\textwidth}\centering
\begin{tcolorbox}[compact, width=\linewidth]
{\bfseries 14}\\[-1pt]
{\bfseries  \scriptsize  Ordinal}\\[-1pt]
{\color{medgray} Education, Age, BMI}
\end{tcolorbox}
\end{minipage}\hfill
\begin{minipage}[t]{0.235\textwidth}\centering
\begin{tcolorbox}[compact, width=\linewidth]
{\bfseries 7}\\[-1pt]
{\bfseries  \scriptsize One-Hot}\\[-1pt]
{\color{medgray} Province, Ethnicity}
\end{tcolorbox}
\end{minipage}\hfill
\begin{minipage}[t]{0.235\textwidth}\centering
\begin{tcolorbox}[compact, width=\linewidth]
{\bfseries 17}\\[-1pt]
{\bfseries  \scriptsize  Binary}\\[-1pt]
{\color{medgray} Yes/No features}
\end{tcolorbox}
\end{minipage}\hfill
\begin{minipage}[t]{0.235\textwidth}\centering
\begin{tcolorbox}[compact, width=\linewidth]
{\bfseries  \scriptsize Data Splitting}\\[-1pt]
{\color{medgray} Training: 80\%}\\[-1pt]
{\color{medgray} Testing: 20\%}
\end{tcolorbox}
\end{minipage}

\vspace{1pt}
\centering{\itshape\color{medgray} Stratified sampling maintains class distribution}
\end{tcolorbox}

\phasearrow

\begin{tcolorbox}[phasebox]
{\bfseries \small Feature Selection Methods}\\[-4pt]

\noindent
\begin{minipage}[t]{0.235\textwidth}\centering
\begin{tcolorbox}[compact, width=\linewidth]
{\bfseries  \scriptsize Chi-Square Test}\\[-1pt]
{\color{medgray} Statistical independence}
\end{tcolorbox}
\end{minipage}\hfill
\begin{minipage}[t]{0.235\textwidth}\centering
\begin{tcolorbox}[compact, width=\linewidth]
{\bfseries \scriptsize Mutual Information}\\[-1pt]
{\color{medgray} Information gain measure}
\end{tcolorbox}
\end{minipage}\hfill
\begin{minipage}[t]{0.235\textwidth}\centering
\begin{tcolorbox}[compact, width=\linewidth]
{\bfseries \scriptsize  Point-Biserial}\\[-1pt]
{\color{medgray} Correlation analysis}
\end{tcolorbox}
\end{minipage}\hfill
\begin{minipage}[t]{0.235\textwidth}\centering
\begin{tcolorbox}[compact, width=\linewidth]
{\bfseries \scriptsize Boruta Algorithm}\\[-1pt]
{\color{medgray} Random Forest based}
\end{tcolorbox}
\end{minipage}

\vspace{2pt}
\centering{\itshape\color{medgray} Consensus: Top 15 features from each method}\\[2pt]

{\bfseries \small Statistical Analysis}\\[0.1pt]
\noindent\begin{minipage}[t]{0.485\textwidth}
\begin{tcolorbox}[innerbox, colframe=darkgray]
{\bfseries  \scriptsize Crude Odds Ratios}\\[-2pt]
{\color{medgray} Univariate associations with custom reference categories}
\end{tcolorbox}
\end{minipage}\hfill
\begin{minipage}[t]{0.485\textwidth}
\begin{tcolorbox}[innerbox, colframe=darkgray]
{\bfseries  \scriptsize Adjusted Odds Ratios}\\[-2pt]
{\color{medgray} Multivariable logistic regression controlling for confounders}
\end{tcolorbox}
\end{minipage}
\textbf{Final: 13 features}
\end{tcolorbox}

\phasearrow

\begin{tcolorbox}[phasebox]
{\bfseries \small Class Imbalance Handling}\\[-6pt]
\begin{tcolorbox}[innerbox, colframe=darkgray]
{\bfseries \small  SMOTE (Synthetic Minority Over-sampling Technique)}\\[-1pt]
{\color{medgray} Integrated within ImbPipeline to prevent data leakage}
\end{tcolorbox}
\end{tcolorbox}

\phasearrow

\begin{tcolorbox}[phasebox]
{\bfseries \small Model Training \& Hyperparameter Tuning}\\[-4pt]

\centering
\begin{tabular}{ccccc}
\begin{tcolorbox}[compact, width=2.25cm, colframe=bordergray, boxrule=0.4pt]\centering\bfseries LR\end{tcolorbox} &
\begin{tcolorbox}[compact, width=2.25cm, colframe=bordergray, boxrule=0.4pt]\centering\bfseries KNN\end{tcolorbox} &
\begin{tcolorbox}[compact, width=2.25cm, colframe=bordergray, boxrule=0.4pt]\centering\bfseries Decision Tree\end{tcolorbox} &
\begin{tcolorbox}[compact, width=2.25cm, colframe=bordergray, boxrule=0.4pt]\centering\bfseries Random Forest\end{tcolorbox} &
\begin{tcolorbox}[compact, width=2.25cm, colframe=bordergray, boxrule=0.4pt]\centering\bfseries XGBoost\end{tcolorbox}
\end{tabular}

\vspace{2pt}

\begin{tabular}{ccccc}
\begin{tcolorbox}[compact, width=2.25cm, colframe=bordergray, boxrule=0.4pt]\centering\bfseries SVM\end{tcolorbox} &
\begin{tcolorbox}[compact, width=2.25cm, colframe=bordergray, boxrule=0.4pt]\centering\bfseries Naive Bayes\end{tcolorbox} &
\begin{tcolorbox}[compact, width=2.25cm, colframe=bordergray, boxrule=0.4pt]\centering\bfseries LDA\end{tcolorbox} &
\begin{tcolorbox}[compact, width=2.25cm, colframe=bordergray, boxrule=0.4pt]\centering\bfseries DNN\end{tcolorbox} &
\begin{tcolorbox}[compact, width=2.25cm, colframe=bordergray, boxrule=0.4pt]\centering\bfseries TabNet\end{tcolorbox}
\end{tabular}

\vspace{2pt}
\begin{tcolorbox}[compact, colframe=darkgray]
{\bfseries\small GridSearchCV Configuration} \quad
{\bfseries CV:} {\color{medgray} RepeatedStratifiedKFold (5 folds $\times$ 3 repeats)} \quad
{\bfseries Scoring:} {\color{medgray} F1-Score} \quad
{\bfseries Search:} {\color{medgray} Exhaustive grid}
\end{tcolorbox}
\end{tcolorbox}

\phasearrow

\begin{tcolorbox}[phasebox]
{\bfseries \small Model Evaluation}\\[-4pt]

\noindent\begin{minipage}[t]{0.36\textwidth}
\begin{tcolorbox}[innerbox, colframe=medgray]
{\bfseries \scriptsize Classification}\\[-1pt]
{\color{medgray} Accuracy $|$ Precision $|$ Recall $|$ F1-Score}
\end{tcolorbox}
\end{minipage}\hfill
\begin{minipage}[t]{0.32\textwidth}
\begin{tcolorbox}[innerbox, colframe=medgray]
{\bfseries  \scriptsize Advanced}\\[-1pt]
{\color{medgray} ROC-AUC $|$ Average Precision}
\end{tcolorbox}
\end{minipage}\hfill
\begin{minipage}[t]{0.28\textwidth}
\begin{tcolorbox}[innerbox, colframe=medgray]
{\bfseries  \scriptsize Validation}\\[-1pt]
{\color{medgray} Train vs Test Score}
\end{tcolorbox}
\end{minipage}
\end{tcolorbox}

\end{minipage}
} 
\caption{Workflow diagram}
\label{fig:workflow_compact}
\end{figure*}

\section*{Methods}
 \subsubsection*{Support Vector Machine} Support Vector Machine seeks an optimal separating hyperplane defined by $f(\mathbf{x})=\mathbf{w}^T\mathbf{x}+b$ that maximizes the margin between classes. The optimization problem minimizes $\frac{1}{2}\|\mathbf{w}\|^2+C\sum_{i=1}^N\xi_i$ subject to $y_i(\mathbf{w}^T\mathbf{x}_i+b)\ge1-\xi_i$, $\xi_i\ge0$. Kernel functions enable nonlinear classification by implicitly mapping data into higher-dimensional spaces \cite{cortes1995support}.
 
 \subsubsection*{Decision Tree} Decision trees are supervised learning models that recursively partition the feature space into disjoint regions using decision rules. Given data $\{(\mathbf{x}_i,y_i)\}_{i=1}^N$, the feature space is divided into regions $\{R_k\}_{k=1}^K$, and predictions are made via $\hat{y}(\mathbf{x})=\sum_{k=1}^K c_k\mathbb{I}(\mathbf{x}\in R_k)$, where $c_k$ denotes the class label assigned to region $R_k$ \cite{breiman2017classification}.

 \subsubsection*{Random Forest} Random Forest is an ensemble learning approach that constructs multiple decision trees using bootstrap samples and random feature selection at each split. The final prediction is obtained by majority voting: $\hat{y}_{\mathrm{RF}}(\mathbf{x})=\operatorname{mode}\{T_1(\mathbf{x}),\dots,T_N(\mathbf{x})\}$. This strategy reduces variance and improves generalization, particularly in high-dimensional and noisy settings \cite{breiman2001random}.
 
 \subsubsection*{K-Nearest Neighbors} K-Nearest Neighbors (KNN) is a non-parametric supervised learning algorithm that classifies samples based on the majority label among their $k$ nearest neighbors in feature space ~\cite{cover1967nearest}. Given a labeled dataset $\{(\mathbf{x}_i,y_i)\}_{i=1}^N$, distances are computed using the Euclidean metric $d(\mathbf{x},\mathbf{x}_i)=\sqrt{\sum_{k=1}^N(x_k-x_{ik})^2}$. The predicted class is obtained as $\hat{y}=\arg\max_c\sum_{\mathbf{x}_i\in\mathcal{N}_k(\mathbf{x})}\mathbb{I}(y_i=c)$.

\subsubsection*{Logistic Regression} Logistic regression  (LR) models the probability of a binary outcome as $p(y=1|\mathbf{x})=\sigma(\mathbf{w}^T\mathbf{x}+b)$, where $\sigma(\cdot)$ is the sigmoid function. Model parameters are estimated by minimizing the negative log-likelihood $L(\mathbf{w},b)=-\sum_{i=1}^N[y^{(i)}\log p^{(i)}+(1-y^{(i)})\log(1-p^{(i)})]$, with optional regularization to prevent overfitting.

\subsubsection*{Deep Neural Network} A Deep Neural Network (DNN) extends artificial neural network architectures by incorporating multiple hidden layers to learn hierarchical representations. For input $\mathbf{x}$, the forward propagation is defined as $\mathbf{h}^{(0)}=\mathbf{x}$ and $\mathbf{h}^{(\ell)}=\sigma^{(\ell)}(\mathbf{W}^{(\ell)}\mathbf{h}^{(\ell-1)}+\mathbf{b}^{(\ell)})$, enabling the extraction of high-level abstract features \cite{rumelhart1986learning}.

 \subsubsection*{TabNet} TabNet is a deep learning architecture designed for tabular data that employs sequential attention to perform instance-wise feature selection. At each decision step, a sparse mask $M_i\in\mathbb{R}^{B\times D}$ is applied to the input features $f$, producing $\tilde{f}_i=M_i\odot f$. The mask is generated via $M_i=\mathrm{sparsemax}(P_{i-1}\odot h_i(a_{i-1}))$, encouraging interpretability through sparse feature usage. A regularization term penalizes excessive feature usage to promote sparsity \cite{arik2021tabnet}.

\subsubsection*{Linear Discriminant Analysis} Linear Discriminant Analysis (LDA) is a supervised dimensionality reduction technique that maximizes class separability by solving the generalized eigenvalue problem $S_B w=\lambda S_W w$, where $S_B$ and $S_W$ denote between-class and within-class scatter matrices. The resulting projections enhance class discrimination and are particularly effective for structured socio-economic and demographic datasets.

 \subsubsection*{Naive Bayes} Naive Bayes  (NB) is a probabilistic classifier based on Bayes’ theorem with the assumption of conditional independence among features. The posterior probability is given by $P(C_k|\mathbf{x})=\frac{P(C_k)\prod_{i=1}^nP(x_i|C_k)}{P(\mathbf{x})}$. Despite its simplicity, NB performs competitively in high-dimensional settings such as text and categorical data classification ~\cite{bishop2006pattern}.

 \subsubsection*{Extreme Gradient Boosting} XGBoost is an ensemble boosting method that builds additive decision trees optimized via gradient descent. The objective function $\mathcal{L}(\theta)=\sum_{i=1}^n L(y^{(i)},f(x^{(i)};\theta))+\Omega(T)+R(w)$ incorporates both loss minimization and regularization to control model complexity, yielding strong predictive performance and robustness to overfitting.

\subsection*{Evaluation Metrics}
Table \ref{metrics} provides an overview of the evaluation metrics used to assess the performance of the proposed models. These metrics facilitate a systematic comparison among the ML and DL models, allowing for a detailed analysis of their strengths and limitations in the classification task of distinguishing between anemic and non-anemic cases within the datasets.

\begin{table}[H]
\centering
\small
\renewcommand{\arraystretch}{1.35}

\begin{tabular}{c c p{4cm} p{8cm}}
\toprule
{S.N.} & {Metrics} & {Formula} & {Description} \\
\midrule

1 & Confusion Matrix 
  & 
  & A matrix summarizing model performance by tallying correctly and incorrectly 
    classified anemic and non-anemic cases as TP, TN, FP, and FN. \\

2 & Accuracy 
  & $\displaystyle \frac{TP + TN}{TP + TN + FP + FN}$
  & It measures the percentage of anemia predictions that the model accurately 
    identifies as either anemic or not anemic. \\

3 & Precision 
  & $\displaystyle \frac{TP}{TP + FP}$
  & It measures the proportion of predicted anemic cases that are truly anemic. \\

4 & Recall 
  & $\displaystyle \frac{TP}{TP + FN}$
  & It measures the model's ability to correctly identify anemic individuals. 
    High recall means low false negatives. \\

5 & F1-score 
  & $\displaystyle \frac{2\,(\text{Precision})(\text{Recall})}{\text{Precision} + \text{Recall}}$
  & The harmonic mean of precision and recall, providing a balanced performance measure. \\

6 & Average Precision 
  & $\displaystyle \sum_{k=1}^{N} P(k)\,[ R(k)-R(k-1) ]$
  & It measures the area under the precision–recall curve, indicating how well the model 
    maintains precision across recall levels. \\

7 & AUC 
  & $\displaystyle \Pr(y_i > y_j)$
  & The probability that the model assigns a higher score to an anemic case than to a 
    non-anemic case across all thresholds. \\

8 & Cohen’s Kappa 
  & $\displaystyle \frac{p_0 - p_e}{1 - p_e}$
  & Measures how much the model's agreement with true labels exceeds random chance, 
    adjusting for class imbalance. \\

\bottomrule
\end{tabular}

\caption{Overview of performance metrics}
\label{metrics}
\end{table}

\section*{Results}
\subsection*{Descriptive Statistics}

\begin{table}[!ht]
\centering
\small
\renewcommand{\arraystretch}{1.3}
\begin{adjustbox}{max width=\textwidth}
\begin{tabular}{llcccccccc}
\toprule
\textbf{Feature} & \textbf{Category} &
\textbf{Anemic} & \textbf{Not Anemic} & \textbf{Total} &
\textbf{Prev. (\%)} &
\textbf{Crude OR} &
\textbf{Adj. OR} & \textbf{Adj. 95\% CI} & \textbf{p} \\
\midrule

\multirow{5}{*}{Child Age}
 & 6--12  & 125 & 41  & 166 & 75.30 & 1.00 &  &  &  \\
 & 13--24 & 213 & 134 & 347 & 61.38 & 0.52 & 0.642 & 0.404--1.021 & 0.061 \\
 & 25--36 & 160 & 226 & 386 & 41.45 & 0.23 & 0.279 & 0.169--0.458 & $<$0.001 \\
 & 37--48 & 154 & 332 & 486 & 31.69 & 0.15 & 0.196 & 0.121--0.317 & $<$0.001 \\
 & 49--59 & 118 & 352 & 470 & 25.11 & 0.11 & 0.137 & 0.084--0.223 & $<$0.001 \\
\midrule

\multirow{2}{*}{Fever}
 & No  & 540 & 835 & 1375 & 39.27 & 1.00 &  &  &  \\
 & Yes & 230 & 250 & 480  & 47.92 & 1.42 & 1.408 & 1.119--1.772 & 0.003 \\
\midrule

\multirow{3}{*}{Household Size}
 & Medium & 415 & 568 & 983 & 42.22 & 1.00 &  &  &  \\
 & Large  & 100 & 104 & 204 & 49.02 & 1.32 & 1.306 & 0.933--1.829 & 0.120 \\
 & Small  & 255 & 413 & 668 & 38.17 & 0.85 & 1.040 & 0.832--1.300 & 0.731 \\
\midrule

\multirow{2}{*}{Mother Anemia}
 & Not anemic & 448 & 776 & 1224 & 36.60 & 1.00 &  &  &  \\
 & Anemic     & 322 & 309 & 631  & 51.03 & 1.81 & 1.621 & 1.299--2.022 & $<$0.001 \\
\midrule

\multirow{2}{*}{Parasite Deworm}
 & No  & 231 & 139 & 370  & 62.43 & 1.00 &  &  &  \\
 & Yes & 539 & 946 & 1485 & 36.30 & 0.34 & 0.650 & 0.479--0.882 & 0.006 \\
\midrule

\multirow{2}{*}{Amenorrhea}
 & No  & 668 & 1001 & 1669 & 40.02 & 1.00 &  &  &  \\
 & Yes & 102 & 84   & 186  & 54.84 & 1.82 & 1.119 & 0.783--1.599 & 0.536 \\
\midrule

\multirow{5}{*}{Ethnicity}
 & Hill Brahmin/Chhetri & 211 & 366 & 577 & 36.57 & 1.00 &  &  &  \\
 & Hill Dalit           & 96  & 152 & 248 & 38.71 & 1.10 & 1.210 & 0.861--1.699 & 0.272 \\
 & Hill Janajati        & 126 & 242 & 368 & 34.24 & 0.90 & 0.982 & 0.703--1.372 & 0.915 \\
 & Other                & 200 & 167 & 367 & 54.50 & 2.08 & 1.974 & 1.391--2.803 & $<$0.001 \\
 & Terai Caste          & 137 & 158 & 295 & 46.44 & 1.50 & 1.449 & 0.961--2.185 & 0.077 \\
\midrule

\multirow{7}{*}{Province}
 & Madhesh             & 176 & 175 & 351 & 50.14 & 1.00 &  &  &  \\
 & Bagmati             & 83  & 130 & 213 & 38.97 & 0.64 & 0.942 & 0.589--1.507 & 0.803 \\
 & Gandaki             & 48  & 105 & 153 & 31.37 & 0.46 & 0.707 & 0.423--1.181 & 0.186 \\
 & Karnali             & 114 & 180 & 294 & 38.78 & 0.63 & 0.902 & 0.569--1.430 & 0.661 \\
 & Koshi               & 96  & 194 & 290 & 33.10 & 0.49 & 0.584 & 0.392--0.869 & 0.008 \\
 & Lumbini             & 132 & 143 & 275 & 48.00 & 0.92 & 0.978 & 0.668--1.432 & 0.910 \\
 & Sudurpashchim       & 121 & 158 & 279 & 43.37 & 0.76 & 0.962 & 0.623--1.485 & 0.861 \\
\midrule

\multirow{2}{*}{Antenatal Care}
 & Adequate ANC   & 716 & 1040 & 1756 & 40.77 & 1.00 &  &  &  \\
 & Inadequate ANC & 54  & 45   & 99   & 54.55 & 1.74 & 0.727 & 0.458--1.154 & 0.177 \\
\midrule

\multirow{2}{*}{Breastfeeding}
 & Yes & 707 & 1022 & 1729 & 40.89 & 1.00 &  &  &  \\
 & No  & 63  & 63   & 126  & 50.00 & 1.45 & 1.265 & 0.828--1.933 & 0.277 \\
\midrule

\multirow{2}{*}{Mother Deworm}
 & Yes & 688 & 1025 & 1713 & 40.16 & 1.00 &  &  &  \\
 & No & 82  & 60   & 142  & 57.75 & 2.04 & 1.06 &  0.71--1.58 & 0.795 \\
\bottomrule
\end{tabular}
\end{adjustbox}
\caption{Factors associated with childhood anemia}
\label{tab:contingency}
\end{table}

Table~\ref{tab:contingency} demonstrates that the overall prevalence of anemia among children aged 6-59 months was 38.5\%. The prevalence decreased with increasing child age. Children aged 6-12 months had the highest prevalence of anemia (75.3\%). Children aged 25-36 months had 72\% lower odds of anemia (AOR = 0.28; 95\% CI: 0.17-0.46; $p < 0.001$). Likewise, those aged 37-48 months (AOR = 0.20; 95\% CI: 0.12-0.32; $p < 0.001$) and 49-59 months (AOR = 0.14; 95\% CI: 0.08-0.22; $p < 0.001$) had significantly lower odds of anemia compared with infants aged 6-12 months. Children who had received deworming tablets had lower odds of anemia (AOR = 0.65; 95\% CI: 0.48--0.88; $p = 0.006$) compared to children who did not receive them. Children having fever in the two weeks preceding the survey had higher odds of anemia (AOR = 1.41; 95\% CI: 1.12-1.77; $p = 0.003$) compared to those without fever.

Some maternal factors were also associated with anemia in children. Children of anemic mothers had 1.62 times higher odds of anemia (AOR = 1.62; 95\% CI: 1.30-2.02; $p < 0.001$) compared to children of non-anemic mothers. Children who were not breastfed had higher odds of anemia (AOR = 1.27; 95\% CI: 0.83-1.93; $p = 0.28$) compared with breastfed children; however, the association was not significant in the multivariate regression analysis. Children born to mothers who had received deworming during pregnancy had lower odds of having anemia compared to those whose mothers did not (AOR = 1.06; 95\% CI: 0.71-1.58; $p = 0.795$).

Variations in the prevalence of anemia among children were observed across demographic factors such as province, household size, and ethnicity. Madhesh province had the highest prevalence of childhood anemia (50.1\%). Children in koshi province had significantly lower odds of anemia compared with those in madhesh (AOR = 0.58; 95\% CI: 0.39--0.87; $p = 0.008$). Similarly, children from the ``Other'' ethnic group had nearly twice the odds of anemia compared with hill brahmin/chhetri children (AOR = 1.97; 95\% CI: 1.39--2.80; $p < 0.001$). Slight differences were observed in anemia prevalence among children from different household sizes; however, the association was not significant in the multivariate regression model.

\subsection*{Model Performance}
Table \ref{results} presents the performance of eight machine learning and two deep learning models. Given the class imbalance and the clinical importance of identifying anemic children, F1-score and recall were prioritized as primary evaluation metrics.

Logistic regression achieved the better overall performance with the highest F1-score (64.9\%) and recall (70.1\%), making it most effective at correctly identifying anemic cases. SVM and DNN demonstrated competitive performance, with F1-scores of 63.6\% and 63.3\%, respectively. SVM also achieved the highest AUC (73.5\%), indicating superior discrimination capability between anemic and non-anemic children. While DNN attained the highest accuracy (70.9\%) and precision (66.4\%), its lower recall (60.4\%) compared to logistic regression indicates it misses more true anemic cases. Random forest and XGBoost achieved identical accuracy (69.0\%) but exhibited lower recall, limiting their effectiveness for this imbalanced classification task. KNN performed poorest across all metrics (F1-score: 57.0\%), while decision tree showed the lowest AUC (66.4\%) with signs of overfitting. The consistency between training and test F1-scores (approximately 60-63\%) across most models indicates appropriate generalization.

\begin{figure*}[!ht]
    \centering
    \includegraphics[width=0.32\textwidth]{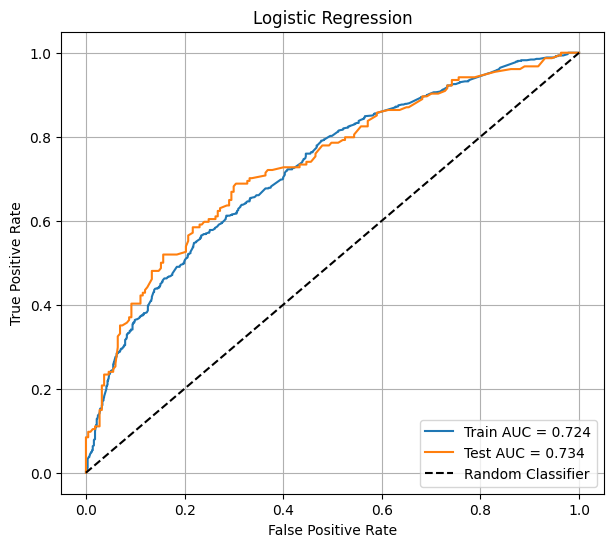}\hfill
    \includegraphics[width=0.32\textwidth]{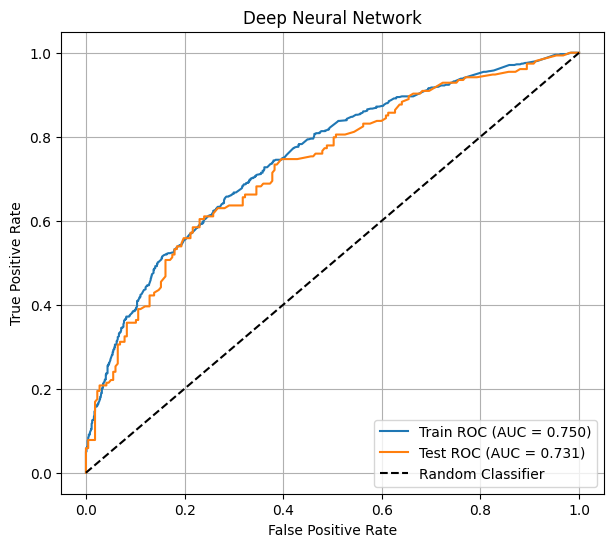}\hfill
    \includegraphics[width=0.32\textwidth]{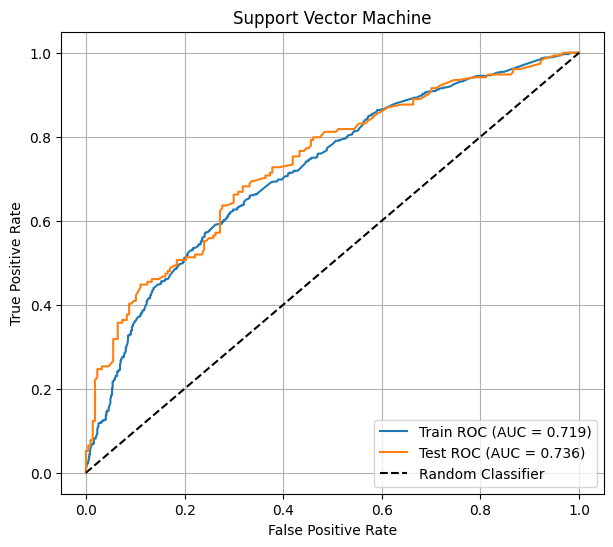}
    \caption{Train and test ROC curves}
    \label{fig:roc_simple}
\end{figure*}

Figure ~\ref{fig:roc_simple} presents the ROC-AUC curves for the three best-performing models: logistic regression, SVM, and DNN. All three models demonstrated comparable discriminatory ability in distinguishing anemic from non-anemic children, with curves substantially above the random classifier diagonal. The close alignment between training and test AUC scores across all models provides strong evidence of robust generalization without overfitting, validating the effectiveness of our cross-validation strategy and SMOTE implementation for handling class imbalance. 

\begin{table}[!ht]
\centering
\small
\renewcommand{\arraystretch}{1.3}
\begin{tabular}{lcccccccccc}
\toprule
 & 
\multicolumn{8}{c}{{Machine Learning}} &
\multicolumn{2}{c}{{Deep Learning}} \\
\cmidrule(lr){2-9} \cmidrule(lr){10-11}
{Metric} &
{LR} & {KNN} & {DT} & {RF} &
{XGB} & {SVM} & {NB} & {LDA} &
{DNN} & {TabNet} \\
\midrule

Accuracy &
0.685 & 0.625 & 0.644 &
0.69 & 0.69 &
0.679 & 0.668 & 0.666 &
\textbf{0.709} & 0.663 \\

Precision &
0.603 & 0.544 & 0.56 & 0.636 & 0.633 &
0.601 & 0.594 & 0.586 &
\textbf{0.664} & 0.582 \\

Recall &
\textbf{0.701} & 0.597 & 0.662 &
0.591 & 0.604 & 0.675 & 0.636 & 0.662 &
0.604 & 0.669 \\

F1 Score &
\textbf{0.649} & 0.57 & 0.607 &
0.613 & 0.618 &
0.636 &
0.614 & 0.622 &
0.633 & 0.622 \\

Average Precision &
\textbf{0.679} & 0.577 & 0.598 &
0.663 & 0.667 &
\textbf{0.684} &
0.676 &
\textbf{0.684} &
0.671 & 0.628 \\

AUC &
\textbf{0.733} & 0.677 & 0.664 &
0.722 & 0.724 &
\textbf{0.736} &
\textbf{0.733} &
\textbf{0.733} &
\textbf{0.731} &
0.709 \\

Cohen's Kappa &
0.365 &
0.239 & 0.286 & 0.355 & 0.357 &
0.351 & 0.324 & 0.324 &
\textbf{0.393} & 0.321 \\

Train Set F1 Score &
0.614 & 0.634 & 0.634 & 0.607 & 0.606 &
0.611 & 0.588 & 0.609 &
0.607 & 0.607 \\
\bottomrule 
\end{tabular}
\caption{Model performance}
\label{results}
\end{table}

\FloatBarrier
\section*{Discussion}
The findings of this study demonstrate that machine learning algorithms can predict the presence of anemia among children under 5 years of age in Nepal, thereby reinforcing the evidence that data-driven modeling approaches can complement traditional epidemiological methods \cite{salma2025evaluating, kebede2023application}. 
Our study showed a clear age-related trend, younger children were more anemic than older ones. The steady decrease in anemia odds with advancing age likely reflects improvements in dietary diversity, maturation of the immune system, and a lower susceptibility to micronutrient deficiencies \cite{WHO_2016_Iron_Supplementation}. 
Likewise, this study demonstrated that childhood anemia was associated with factors such as parasitic deworming, maternal deworming, province, recent fever, and other ethnic groups; consistent with previous studies \cite{Upadhyay2025_anemiaBayesian, Chowdhury2020,kuniyoshi2025maternal}.
Given the class imbalance in our dataset, we prioritized F1-score and recall as primary evaluation metrics. F1-score provides a harmonic balance between precision and recall, while recall specifically measures the model's ability to correctly identify anemic cases among all truly anemic children, a critical consideration for public health screening where missing anemic cases have serious clinical implications. The comparatively lower precision score in our study highlights the need for confirmatory hemoglobin testing for false positive cases, as also highlighted by studies using alternative methods of screening \cite{HemoglobinDetection2025, ClinicalSignsAnemia2025}.

Our results demonstrate that both machine learning and deep learning approaches achieve comparable predictive performance. Logistic regression emerged as the best performing model, achieving the highest F1-score (64.9\%) and recall (70.1\%), followed closely by SVM (F1: 63.6\%, recall: 67.5\%) and DNN (F1: 63.3\%, recall: 60.4\%). The superior performance of logistic regression can be attributed to its convex optimization, ensuring convergence to a global minimum, and its limited parameter space, reducing the risk of overfitting \cite{hastie2009elements, goodfellow2016deep}. It has also been a preferred method  in traditional epidemiological studies to determine the association of variables predicting anemia across low-and middle-income countries in DHS data\cite{Chowdhury2020}.  The close alignment between training F1-score (61.4\%) and test F1-score (64.9\%) indicates stable generalization without overfitting. 

DNN demonstrated competitive performance with the highest accuracy (70.9\%) and precision (66.4\%) as it can model complex nonlinear decision boundaries and feature interactions \cite{lecun2015deeplearning,Ye2024ACL}. The relatively small feature space (13 features) in our study mitigates the curse of dimensionality and supports generalization, as evidenced by consistent training (60.7\%) and test (63.3\%) F1-scores. SMOTE was employed to address class imbalance by synthetically augmenting the minority class, helping models learn more balanced decision boundaries and improving sensitivity to anemic cases \cite{chawla2002smote, he2009learning}. Random Forest, XGBoost, SVM, and LDA showed consistently strong performance across multiple metrics (accuracy: 66.6-69.0\%, AUC: 72.2-73.5\%), indicating that multiple robust learners can generalize effectively with the given feature representation.

KNN exhibited the poorest performance (F1: 57.0\%, recall: 59.7\%), likely because SMOTE can distort local neighborhood structures that distance-based learners rely upon \cite{chawla2002smote}, and predominantly categorical features create distance metric ambiguity \cite{hu2016distanceknn, alfeilat2019distanceknn}. TabNet also underperformed (F1: 62.2\%), as its architecture designed for complex feature interactions provides limited advantage with small sample sizes ($n=1,855$) and predominantly categorical data \cite{arik2021tabnet, shwartz-ziv2022tabular}.

The moderate performance levels observed across our models (approximately 60-70\% across key metrics) are consistent with expectations when using indirect sociodemographic and household-level features rather than direct clinical measurements. Studies using clinical and laboratory data have reported substantially higher accuracy (>96\%) for anemia classification, as these sources provide direct physiological indicators of hemoglobin levels and iron status \cite{mahmood2025basrah,article1}. In contrast, our approach relies on proxy variables such as child age, recent fever, household size, ethnicity, and deworming, which capture upstream social determinants rather than immediate biological markers. This fundamental difference in feature types establishes an expected performance ceiling for sociodemographic models. However, non-invasive prediction based methods like this may provide a more economical way of screening for anemia in low-and middle-income countries like Nepal where screening is scarce due to resource limitations  \cite{Sriram2025}. Non-invasive approaches occupy an intermediate position: nail image-based methods achieved approximately 70\% accuracy \cite{10.3389/fdata.2025.1557600}, elastic net logistic regression using hematological features achieved 97\% accuracy differentiating iron deficiency anemia from $\beta$-thalassemia trait \cite{mahmood2025basrah}, and facial image-based screening demonstrated 84\% accuracy with 93\% sensitivity \cite{10.3389/fpubh.2022.964385}.

Machine learning studies on DHS data show variable performance depending on sample size, features, and class separability across different countries. A Bangladesh DHS 2011 study ($n=2,013$ with 24 variables) reported random forest performance of 68.5\% accuracy, 70.7\% recall, and 66.4\% precision for childhood anemia \cite{khan_chowdhury_islam_raheem_2022}, comparable to our results. Ethiopian DHS analysis ($n=8,482$ with 13 features) identified logistic regression as the best-performing method with 66\% accuracy and 82\% recall \cite{10.1371/journal.pone.0300172}, closely aligned with our findings. These comparable performance ranges across diverse DHS-based studies reinforce that sociodemographic features consistently yield models in the 60-75\% accuracy range, reflecting the inherent predictive limitations of indirect indicators.

Our NDHS-based analysis demonstrates competitive predictive performance despite a smaller sample ($n=1,855$) and fewer features (13 features). While previous studies using NDHS focused on spatial patterns and determinants of anemia \cite{sharma2022spatial,kuniyoshi2025maternal}, our study uniquely applies machine learning for prediction as well. The findings suggest the feasibility of developing community-based risk scores using accessible features: child age, recent fever, maternal anemia, household size, province, deworming status, amenorrhea, antenatal care adequacy, and breastfeeding practices consistent with studies using DHS data sets \cite{Chowdhury2020, kuniyoshi2025maternal, kebede2023application}. Although these models cannot replace laboratory-confirmed diagnoses, they provide valuable tools for initial screening and resource allocation in settings where routine hemoglobin testing is unavailable \cite{Sriram2025}. Children identified as high-risk could be referred for hemoglobin testing followed by targeted interventions including iron supplementation and fever prevention. Our results emphasize the importance of integrated maternal-child health approaches, as maternal factors (anemia status, deworming during pregnancy, adequate antenatal care) substantially influence childhood anemia risk \cite{Chowdhury2020, kuniyoshi2025maternal}.

\emph{Limitations}

Our analysis retained 1,855 observations after excluding missing values, potentially limiting generalizability. The moderate performance (F1-score: 64.9\%) reflects inherent constraints of sociodemographic predictors, these models serve as screening tools requiring confirmatory hemoglobin testing, not diagnostic replacements. The cross-sectional design precludes causal inference, and SMOTE-generated synthetic samples may introduce artificial patterns.

\section*{Conclusion}
Our study provides a methodologically distinct contribution by systematically benchmarking ten machine learning and deep learning algorithms for anemia prediction using NDHS 2022 data. We integrated 13 features spanning child demographics, maternal health, and household characteristics, employing stratified splitting, SMOTE for class imbalance, and rigorous hyperparameter optimization through GridSearchCV with repeated stratified k-fold cross-validation. Logistic regression achieved optimal performance with an F1-score of 64.9\% and recall of 70.1\%. The consistent performance range of 60-70\% across models reflects the predictive capacity of sociodemographic proxies versus direct clinical measurements, remaining competitive with similar DHS studies. This framework enables early risk stratification using readily available information from routine health surveys, facilitating targeted screening where universal hemoglobin testing is infeasible. Community health workers could apply risk scores to identify children requiring confirmatory testing, optimizing resource allocation for iron supplementation and nutritional interventions. Integration with maternal-child health services could support data-driven decision-making for antenatal care improvement and household-level interventions \cite{Zemariam2024, usanzineza2024prevalence}.

\emph{Future Work}

Future research should incorporate multiple NDHS waves to increase sample size and enable temporal validation. Integrating supplementary data such as dietary diversity, micronutrient compliance, helminth burden, and geographic information systems could enhance accuracy. Developing interpretable risk scores for mobile health platforms would facilitate community implementation, while expanding to multi-class severity prediction would enable differentiated care pathways and targeted interventions.

\section*{Conflict of Interest}
The authors declare no conflicts of interest.

\section*{Data and Code Availability}
The datasets analyzed in this study are publicly accessible through the Demographic and Health Surveys (DHS) repository (\url{https://dhsprogram.com/data/}). Researchers interested in the code used for this analysis may obtain it from the corresponding author upon request.

\bibliographystyle{plain}
\bibliography{ref}
\end{document}